\documentclass[letterpaper, 10 pt, conference]{ieeeconf}  

\IEEEoverridecommandlockouts                              
\usepackage{graphics}
\usepackage{makeidx}  
\usepackage{amsmath}
\usepackage{amssymb}
\usepackage{latexsym}
\usepackage{epsfig}
\usepackage{verbatim}
\usepackage{psfrag}
\usepackage{graphicx}
\usepackage{subfigure}
\usepackage{amssymb}
\usepackage{bm}
\renewcommand{\vec}[1]{\bm{#1}}
\newcommand{\mat}[1]{\bm{#1}}

\title{Implicit Gradient Neural Networks with a Positive-Definite Mass Matrix for Online Linear Equations Solving
}

\author{Ke Chen
}

\begin{document}

\maketitle
\thispagestyle{empty}
\pagestyle{empty}

\begin{abstract}

Motivated by the advantages achieved by implicit analogue net for solving online linear equations, a novel implicit neural model is designed based on conventional explicit gradient neural networks in this letter by introducing a positive-definite mass matrix. In addition to taking the advantages of the implicit neural dynamics, the proposed implicit gradient neural networks can still achieve globally exponential convergence to the unique theoretical solution of linear equations and also global stability even under no-solution and multi-solution situations. Simulative results verify theoretical convergence analysis on the proposed neural dynamics.

\end{abstract}

\section{Introduction}\label{sec.introd}

Researchers in science and engineering have paid their continuous concern on addressing the problem of online linear time-invariant equations, which is formulated as:
\begin{equation}
\mat{A}\vec{x}(t)-\vec{b}=\vec{0}~~\text{or}~~\mat{A}\vec{x}(t)=\vec{b},\label{eqn.le}
\end{equation}
where $\mat{A}\in \mathbb{R}^{n\times n}$ and $\vec{b} \in \mathbb{R}^{n}$ are  coefficients, and the unknown vector $ \vec{x}(t)\in \mathbb{R}^{n}$ denotes the real-time solution with time $t$ evolves. 
When coefficient $\mat{A}$ is non-singular, $\vec{x}(t)$ has an unique theoretical solution $\vec{x}^*=\mat{A}^{-1}\vec{b}$; when coefficient $\mat{A}$ is singular, linear equations (\ref{eqn.le}) can have no solution or multiple solutions.
A number of related problems such as matrix inversion \cite{YZ09AMC,Chen2013amc,Chen2016amc,Chen2007icic,zyck09tac}, quadratic programming/minimisation \cite{Chen2016bics,Chen2008icic,zy2011neurocomputing}, Sylvester equations \cite{Chen2016ipl}, Lyapunov matrix equation \cite{yi2011ipl,chen2013ipl} and linear matrix equations $\mat{A}\mat{X}(t)\mat{B} = \mat{C}$ \cite{Chen2008icic2,Chen2008icit}
can be transformed into linear equations (\ref{eqn.le}) via vectorisation and Kronecker product \cite{Chen2007icic}. Such a problem is widely encountered in other fields of science and engineering such as ridge regression in machine learning \cite{Chen17pr,Chen12bmvc,Chen13cvpr,Chen16tits}, signal processing \cite{ck2017tim}, optical flow in computer vision \cite{Beauchemin95CSUR,Baker2011IJCV}, and robotic inverse kinematics \cite{Chen2008ISSCAA,zy2011RCIM}. 

To address such a problem, the existing algorithms generally can fall into three categories: direct solution based on matrix inversion, serial-processing numerical algorithms, and the parallel-processing algorithms. 
Evidently, either inverse based direct solution or numerical algorithms are less favourable to cope with large-scale or real-time problems because of its high computational cost (typically $O(n^3)$ operations, i.e. the cube of the dimensionality $n$ of coefficient-matrix $\mat{A}$). 
For less computational efficiency, some $O(n^2)$-operation algorithms are proposed in
\cite{YZWELDJL05cdc,WELYZ07cssc}.

Analogue recurrent neural networks are thus proposed and analysed owing to its high computational efficiency, rich information distributed in neurons and the feasibility to implement the model into circuits \cite{wj92el,rz94el,zyck08el,ck12el,cyzy08el,ck13el}. 
A kind of gradient-based explicit dynamical network (termed as gradient neural networks) was proposed for real-time linear equations solving \cite{wj92el,rz94el,zyck08el,ck12el}. 
Gradient neural networks in the form of an explicit system ($\dot{\vec{x}}(t)=\cdots$) can be obtained based on the negative gradient of a scalar-valued energy function $\|\mat{A}\vec{x}(t)-\vec{b}\|_2^2/2\in \mathbb{R}$ as follows \cite{wj92el,rz94el,zyck08el,ck12el}:
\begin{equation}
\dot{\vec{x}}(t) = - \gamma \mat{A}^T( \mat{A}\vec{x}(t) - \vec{b}),\label{eqn.wnn}
\end{equation}
where design parameter $\gamma>0$ is utilised to control the speed of convergence. 
Note that, $\gamma$ should be set as large as possible and we select $\gamma =1000$ for illustrative purpose in the simulative verification in Section \ref{sec.simulation}.

With the advantages of adopting an implicit-dynamics form (i.e., $\mat{M}\dot{\vec{x}}(t)$ $=\cdots$ with the mass matrix $\mat{M}$), Zhang neural networks for solving online linear equations can be achieved by time derivative of a vector-formed error function $\mat{A}\vec{x}(t)-\vec{b} \in \mathbb{R}^n$ \cite{cyzy08el} as follows: 
\begin{equation}
\mat{A}\dot{\vec{x}}(t) = - \gamma ( \mat{A}\vec{x}(t) - \vec{b}).\label{eqn.znn}
\end{equation}

Recently, a novel implicit neural mode (termed as improved Zhang neural networks) \cite{ck13el} was proposed to achieve superior convergence performance in comparison with the explicit gradient net (\ref{eqn.wnn}) and implicit Zhang net (\ref{eqn.znn}), which can be depicted as follows:
\begin{equation}
\mat{A}\dot{\vec{x}}(t) = - \gamma (\mat{A}\mat{A}^T + \mat{I})(\mat{A}\vec{x}(t) - \vec{b}),\label{eqn.cnn}
\end{equation} 
where $\mat{I} \in \mathbb{R}^{n \times n}$ is the identity matrix. 

Compared to explicit dynamic models (e.g., gradient neural networks (\ref{eqn.wnn})), implicit dynamic systems (e.g., conventional Zhang model (\ref{eqn.znn}) and improved Zhang model (\ref{eqn.cnn})) have higher abilities in representing dynamic systems because of preserving physical parameters in the coefficient matrices, e.g., $\mat{A}$ on the left-hand side of (\ref{eqn.znn}) and (\ref{eqn.cnn}) \cite{cyzy08el}. 
In this sense, owing to the advantages of implicit systems,  Zhang models (\ref{eqn.znn}) and (\ref{eqn.cnn})
can be much superior to gradient neural networks (\ref{eqn.wnn}) in the form of explicit dynamics.
However, in view of mass matrix $\mat{M} = \mat{A}$, the existing implicit neural systems (\ref{eqn.znn}) and (\ref{eqn.cnn})) become differential-algebraic-equations (can be formulated as constrained ordinary-differential-equations and abbreviated as DAE), which cannot be completely solvable and might become an initial state problem \cite{DAE}, when coefficient matrix $\mat{A}$ is not invertible (i.e., the mass matrix is singular).

This letter proposes a novel model (termed as implicit gradient neural networks), which can both keep the advantages of implicit neural dynamics about rich representation of neural dynamics for circuit realisation \cite{cyzy08el} and also overcome the drawbacks of the existing implicit neural models (\ref{eqn.znn}) and (\ref{eqn.cnn}) on the difficulty in coping with the problem of singular mass matrices. 
Owing to adopting a positive-definite mass matrix, our model can still achieve global stability even when coefficient matrix $\mat{A}$ is singular (i.e., no-solution and multi-solution cases). 
Computer simulation results can demonstrate theoretical analysis on the proposed analogue network for online solution of simultaneous linear equations (\ref{eqn.le}).

The contributions and novelties of this letter is as follows.
\begin{itemize}
\item To the best of our knowledge, a positive-definite mass matrix is for the first time introduced to the implicit neural dynamics, which can effectively cope with singular coefficient matrix.
\item The proposed model can achieve global exponential convergence when linear equations have a unique solution, while it can still be global stable even for no-solution and multi-solution cases. 
\end{itemize}

\section{Model Formulation}

Based on the explicit gradient neural model (\ref{eqn.wnn}), the following design procedure is adopted to obtain the proposed implicit gradient neural networks.
\begin{itemize}
\item We firstly define a scalar-valued norm-based energy function $\|\mat{A}\vec{x}(t)-\vec{b}\|_2^2/2$. Such an object function can reach its minimum point if and only if the solution $\vec{x}(t)$ of linear equations (\ref{eqn.le}) is equal to its theoretical solution $\vec{x}^* = \mat{A}^{-1}\vec{b}$.
\item Secondly, our model is designed to evolve along the negative gradient of this energy function $\|\mat{A}\vec{x}(t)-\vec{b}\|_2^2/2$ until the minimum is reached \cite{zy09automatica}:
$-\frac{\partial (\|\mat{A}\vec{x}(t)-\vec{b}\|_2^2/2)}{\partial \vec{x}(t)} = -\mat{A}^T\mat{A}\vec{x}(t)+\mat{A}^T\vec{b}$.
\item Finally, different from explicit gradient model (\ref{eqn.wnn}) adopting an identity mass matrix $\mat{I}$, a positive-definite mass matrix $\mat{A}^T\mat{A}+\mat{I}$ is introduced into our implicit dynamical model. 
Consequently, we can obtain the following implicit gradient neural networks:
\begin{equation}
(\mat{A}^T\mat{A}+\mat{I})\dot{\mathbf{x}}(t) = - \gamma \mat{A}^T(\mat{A}\vec{x}(t) - \vec{b}).\label{eqn.iwnn}
\end{equation}
\end{itemize}
Evidently, our model (\ref{eqn.iwnn}) remains ordinary-differential-equations (ODE) owing to the usage of the positive-definite mass matrix $\mat{A}^T\mat{A}+\mat{I}$ even when $\mat{A}$ is singular
and still achieve global stability for no-solution and multi-solution cases. In addition, for hardware realisation, the proposed implicit neural dynamics (\ref{eqn.iwnn}) has higher representing capability of dynamic systems in comparison with explicit gradient neural networks (\ref{eqn.wnn}).


\section{Global Exponential Convergence}

We analyse the global exponential convergence of the proposed implicit gradient neural networks (\ref{eqn.iwnn}) when linear equations (\ref{eqn.le}) have a unique solution, i.e., coefficient $\mat{A}$ in linear equations (\ref{eqn.le}) is non-singular.

\vspace{0.1cm} \noindent \textit{Theorem 1:} Given a non-singular coefficient matrix $\mat{A} \in \mathbb{R}^{n \times n}$ and a coefficient vector $\vec{b}\in \mathbb{R}^n$, the state solution $\vec{x}(t)\in \mathbb{R}^n$ of implicit gradient neural networks (\ref{eqn.iwnn}) can achieve globally exponential convergence to a unique exact solution $\vec{x}^*=\mat{A}^{-1}\vec{b}$ of linear equations (\ref{eqn.le}). $\hfill\Box$

\vspace{0.1cm} \noindent\textit{Proof:} Given the exact solution $\vec{x}^*$,
we can define $\vec{e}(t):=\vec{x}(t)-\vec{x}^*\in \mathbb{R}^n$ as
the solution error by implicit gradient neural networks (\ref{eqn.iwnn}). 
Consequently, time derivation of $\vec{e}(t)$ can be as the following:
$\dot{\vec{e}}(t)=\dot{\vec{x}}(t)$. Let us substitute
$\vec{e}(t)$ as well as $\dot{\vec{e}}(t)$ to implicit model
(\ref{eqn.iwnn}) and we can get the following equation:
\begin{equation}
(\mat{A}^T\mat{A}+\mat{I})\dot{\vec{e}}(t) = - \gamma
\mat{A}^T\mat{A}\vec{e}(t). \label{eqn.iwnn_err}
\end{equation}
Let us define a positive-definite Lyapunov function candidate
$\varphi(t)=\|(\mat{A}^T\mat{A}+\mat{I})\vec{e}(t)\|_2^2/2\geqslant 0$. 
It is evident that $\varphi(t)$ is positive definite in the sense that $\varphi(t)>0$ for any $\vec{e}(t)\neq \vec{0}$, and $\varphi(t)=0$ only for $\vec{e}(t)=\vec{0}$ (i.e., $\vec{x}(t)$ is equal to the exact solution $\vec{x}^*=\mat{A}^{-1}\vec{b}$ of linear equations $\mat{A}\vec{x}(t)=\vec{b}$). 
As a result,
$\dot{\varphi}(t)$ along the state trajectory of the reformulated implicit model (\ref{eqn.iwnn_err}) using the solution error $\vec{e}(t)$ is written as follows:
$\dot{\varphi}(t)
=\vec{e}^T(t)(\mat{A}^T\mat{A}+\mat{I})^T(\mat{A}^T\mat{A}+\mat{I})\dot{\vec{e}}
(t)
\leqslant -\gamma \alpha \beta \|\vec{e}(t)\|_2^2 \leqslant 0$,
where $\alpha>0$ and $\beta>1$ denote the minimum eigenvalue of
$\mat{A}^T\mat{A}$ and $\mat{A}^T\mat{A} + \mat{I}$ respectively. In the light of Lyapunov stability theory \cite{zyck08el,ck13el}, the proposed implicit neural model (\ref{eqn.iwnn_err}) thus reaches 
globally asymptotic stability. For more desirable exponential
convergence \cite{zyck08el,ck13el}, $\dot{\varphi}(t)$ can be reformulated
as
\begin{equation*}
\begin{split}
\dot{\varphi}(t) &\leqslant - \gamma \vec{e}^T(\mat{A}^T\mat{A}+\mat{I})^T \mat{A}^T\mat{A}\vec{e}(t)\\
&\leqslant - \gamma \vec{e}^T\mat{A}^T\mat{A} \mat{A}^T\mat{A}\vec{e}(t)  \\
& ~~~~- \gamma \vec{e}^T\mat{A}^T\mat{A}^T
\vec{e}(t) \\
&\leqslant - \gamma \|\mat{A}^T\mat{A}\vec{e}(t)\|_2^2 - \gamma \alpha \|\vec{e}(t)\|_2^2.
\end{split}
\end{equation*}
The above formulation can fall into two sub-situation: 1) $0<\alpha<1$ and 2) $\alpha\geqslant 1$. 
\begin{itemize}
\item For the former situation $0<\alpha<1$, $\dot{\varphi}(t) \leqslant -\gamma \alpha (\|\mat{A}^T\mat{A}\vec{e}(t)\|_2^2 + \|\vec{e}(t)\|_2^2) \leqslant -2\gamma \alpha {\varphi}(t) $ and thus exponential convergence rate is $\gamma \alpha$.
\item For the latter situation $\alpha\geqslant 1$, $\dot{\varphi}(t)\leqslant -\gamma (\|\mat{A}^T\mat{A}\vec{e}(t)\|_2^2 + \|\vec{e}(t)\|_2^2) \leqslant -2\gamma {\varphi}(t) $ and thus exponential convergence rate of (\ref{eqn.iwnn}) to $\vec{x}^*$ is $\gamma$. 
\end{itemize}
Proof of Theorem 1 is thus completed. $\hfill\Box$
 
\section{Global Stability}
\begin{figure}[t]
\psfrag{t}[c][c][1]{time $t$, ms~~~} %
\psfrag{ax-b}[c][c][1]{~~~~~~~~$\|\mat{A}\vec{x}(t)-\vec{b}\|_2$} %
\centering {%
{\includegraphics[width=0.98\columnwidth]{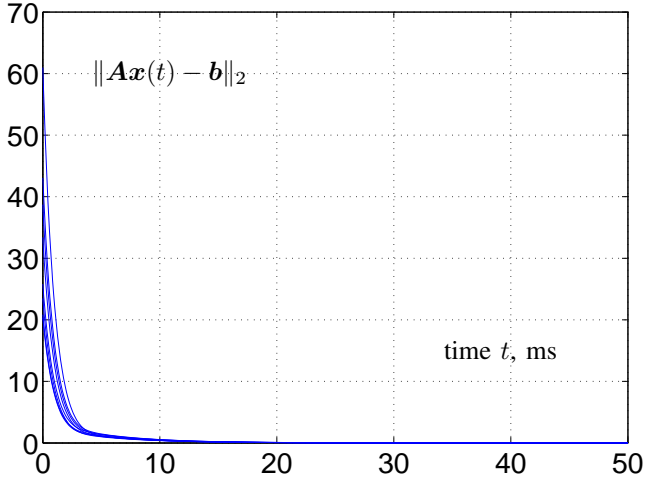}}
}
\caption{Convergence of residual error of the proposed implicit gradient neural networks (\ref{eqn.iwnn}) with $\gamma=1000$ from six initial states randomly-selected within $[-2,2]$. %
}\label{fig.1}
\end{figure}

Global stability of the proposed implicit dynamical model (\ref{eqn.iwnn}) can also be achieved when coefficient $\mat{A}$ is singular as follows.

\vspace{0.1cm} \noindent
\textit{Theorem 2:} Given a singular coefficient matrix $\mat{A} \in \mathbb{R}^{n \times n}$ and a coefficient vector $\vec{b}\in \mathbb{R}^n$, the state vector $\vec{x}(t)\in \mathbb{R}^n$ of implicit gradient neural networks (\ref{eqn.iwnn}) can still be globally stable. $\hfill\Box$

\vspace{0.1cm} \noindent\textit{Proof:} Without the unique solution for singular matrix $\mat{A}$, the following Lyapunov function candidate
$\varphi(t)=\|\mat{A}^T\mat{A}\vec{x}(t)-\mat{A}^T\vec{b}\|_2^2/2 + \|\mat{A}\vec{x}(t)-\vec{b}\|_2^2/2\geqslant 0$ is considered. Then, along the state trajectory of implicit gradient neural networks (\ref{eqn.iwnn}), time derivative of $\varphi(t)$ can be derived as:
\begin{equation*}
\begin{split}
\dot{\varphi}(t) &= (\mat{A}\vec{x}(t)-\vec{b})^T\mat{A}\mat{A}^T\mat{A}\dot{\vec{x}}(t) \\
 &~~~~+ (\mat{A}\vec{x}(t)-\vec{b})^T\mat{A}\dot{\vec{x}}(t)\\
&\leqslant - \gamma (\mat{A}\vec{x}(t)-\vec{b})^T\mat{A}\mat{A}^T(\mat{A}\vec{x}(t) - \vec{b}) \\
&= -\gamma\| \mat{A}^T(\mat{A}\vec{x}(t) - \vec{b}) \|_2^2\leqslant 0.
\end{split}
\end{equation*}
According to Lyapunov stability theory \cite{zyck08el,ck13el}, the implicit gradient neural networks (\ref{eqn.iwnn}) is globally stable. $\hfill\Box$

\section{Simulative Verification}\label{sec.simulation}
\begin{figure*}[t]
\psfrag{t}[c][c][1]{time $t$, ms} %
\psfrag{ax-b}[c][c][1]{~~~~~~~~$\|\mat{A}\vec{x}(t)-\vec{b}\|_2$} %
\centering{ %
\subfigure[no-solution case]{\includegraphics[width=0.98\columnwidth]{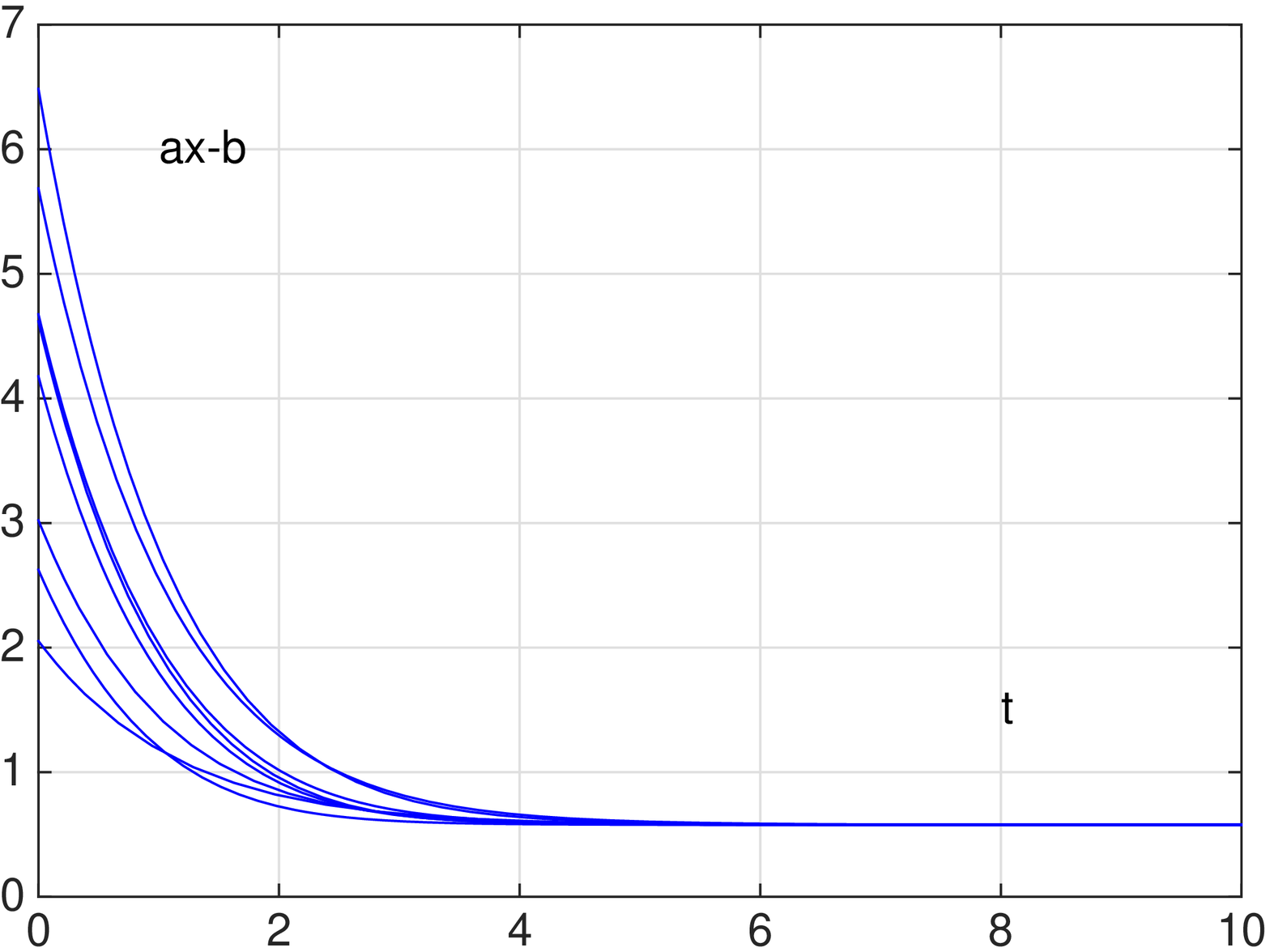}}~~
\subfigure[multi-solution case]{\includegraphics[width=0.98\columnwidth]{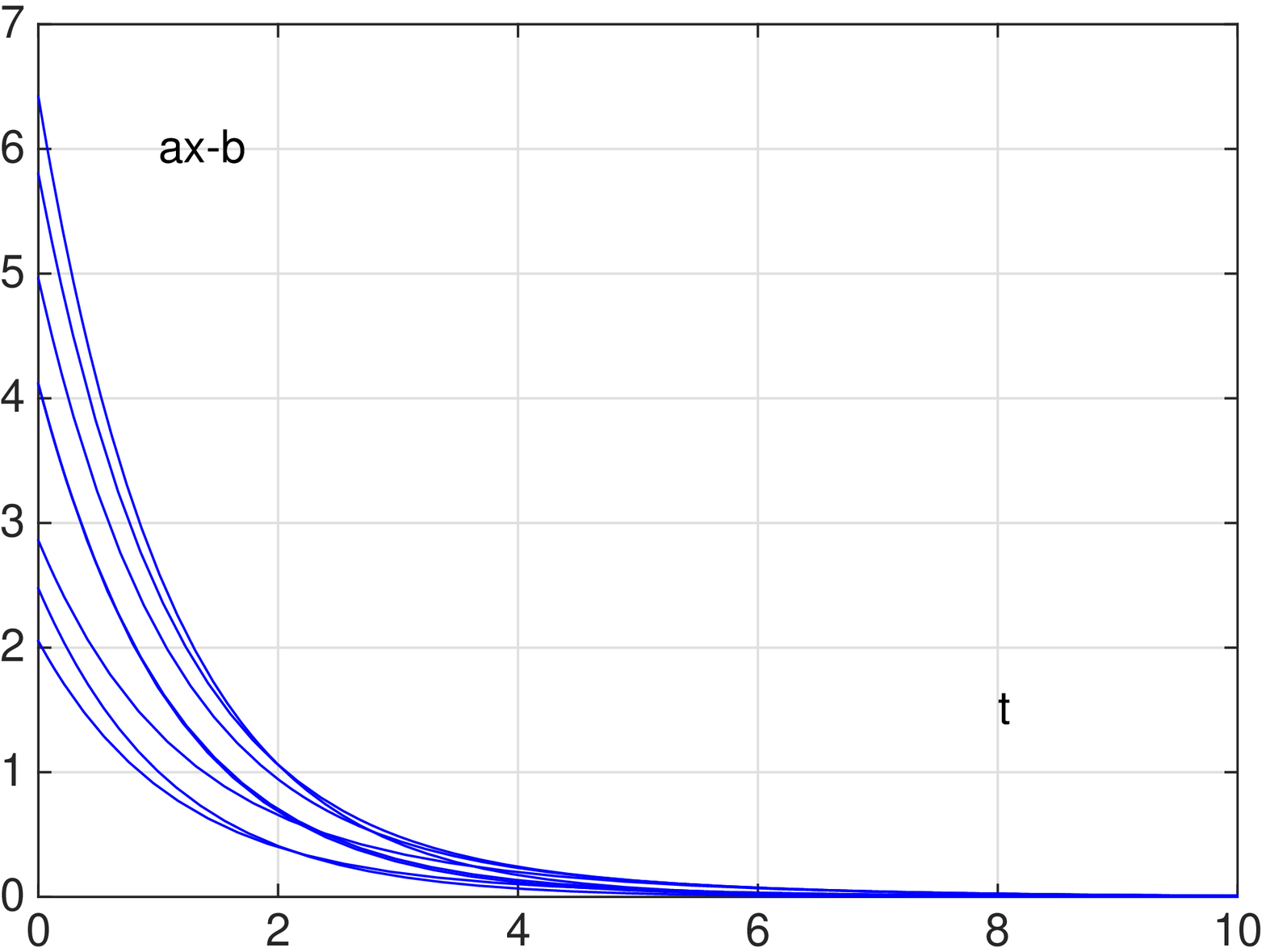}}
}
\caption{Convergence of residual error by the proposed implicit dynamical model (\ref{eqn.iwnn}) implying global stability for both no-solution case and multiple-solution case with $\gamma=1000$
}\label{fig.2} %
\end{figure*}

For verifying the aforementioned analysis, we consider the same illustrative example in \cite{wj92el,zyck08el,ck13el}.
Related to the non-singular coefficient matrix $\mat{A}$, the minimal eigenvalue $\alpha$ of $\mat{A}^T\mat{A}$ is $0.2345$, which falls into the first situation $0<\alpha<1$. 
As global exponential convergence rate $\alpha\gamma$ of the implicit neural networks (\ref{eqn.iwnn}) can be $234.5$ with the design parameter $\gamma = 1000$ according to the analysis of Theorem 1,
convergence time to a tiny residual error $\exp(-7)$ is 29.9 $\text{ms}$. 
Simulative result about the residual error $\|\mat{A}\vec{x}(t)-\vec{b}\|_2$ is shown in Figure \ref{fig.1} to demonstrate globally exponential convergence of the proposed implicit dynamical model (\ref{eqn.iwnn}), from six initial states randomly-selected within $[-2,2]$. 

To show the global stability
of the proposed model (\ref{eqn.iwnn}) in Theorem 2, we consider the singular coefficient matrix $\mat{A}$ (i.e., $\alpha = 0$) and two coefficient vector $\vec{b}$ corresponding to no-solution and multi-solution cases respectively: 
\begin{equation*}
\begin{split}  
\mat{A} = 
\begin{bmatrix}
1  & -1  & 0\\
-1 & 2   & 1\\
0  & 1   & 1\\
\end{bmatrix},
\vec{b}^{(1)} = 
\begin{bmatrix}
1  \\
1  \\
1  \\
\end{bmatrix},
\vec{b}^{(2)} = 
\begin{bmatrix}
0  \\
1  \\
1  \\
\end{bmatrix},
\end{split}
\end{equation*}
where, if $\vec{b} = \vec{b}^{(1)}$ linear equations $\mat{A}\vec{x}(t)=\vec{b}$ has no theoretical solution; if $\vec{b} = \vec{b}^{(2)}$ there is multiple solutions. 
In Figure \ref{fig.2}, we visualise simulation results about the residual error $\|\mat{A}\vec{x}(t)-\vec{b}\|_2$ for both no-solution and multi-solution cases, which can verify theoretical analysis on the implicit dynamical model (\ref{eqn.iwnn}) in Theorem 2.


\section{Conclusion}

A novel implicit gradient  neural network for real-time solution of linear equations is proposed and analysed under different situations. 
Global exponential convergence and stability on implicit dynamical models can be achieved owing to the introduction of a positive-definite mass matrix. 
Simulation results substantiate theoretical analysis on the proposed neural networks. 

%

\end{document}